\documentclass{article}  

\usepackage{amsmath,amssymb}
\usepackage{graphicx}
\usepackage{subcaption}
\usepackage{mathtools}
\usepackage{algorithm}
\usepackage{hyperref}
\usepackage{xcolor}
\usepackage{siunitx}

\begin{document}

\title{Computing Funnels Using Numerical Optimization Based Falsifiers\thanks{This work was supported by the project GA21-09458S of the Czech Science
Foundation GA {\v C}R and institutional support RVO:67985807.}}

\author{Ji{\v r}{\'\i} Fejlek$^{1,2}$\footnote{ORCID: 0000-0002-9498-3460} \;and Stefan Ratschan$^{1}$\footnote{ORCID: 0000-0003-1710-1513}\\ 
$^1$The Czech Academy of Sciences, Institute of Computer Science\\
$^2$Czech Technical University in Prague, Faculty of Nuclear Sciences and Physical Engineering}

\maketitle
\thispagestyle{empty}
\pagestyle{empty}

\begin{abstract}
In this paper, we present an~algorithm that computes funnels along trajectories of systems of ordinary differential equations. A funnel is a time-varying set of states containing the given trajectory, for which the evolution from within the set at any given time stays in the funnel. Hence it generalizes the behavior of single trajectories to sets around them, which is an important task, for example, in robot motion planning.

In contrast to approaches based on sum-of-squares programming, which poorly scale to high dimensions,
our approach is based on falsification and tackles the funnel computation task directly, through numerical optimization. This approach computes accurate funnel estimates far more efficiently and leaves formal verification to the end, outside all funnel size optimization loops. 
\end{abstract}

\section{Introduction}

An important task in robotics motion planning is to follow a given trajectory into some target set \cite{review1,review2,review3}. Especially, numerous path planning algorithms~\cite{funnels1,LQRtrees1,funnels2} rely on this task. This involves first designing a controller that follows this trajectory~\cite{funnels3}, and then determining a neighbourhood of the trajectory (a \emph{funnel}~\cite{SOS_funnels}) where the controller fulfils its goal of reaching a given target set \cite{funnels4,funnels5}. In this paper, we present an efficient method for this second task.

In the literature~\cite{LQRtrees1,SOS_funnels,SOS6}, funnel construction is usually based on sum-of-squares programming (SOS)~\cite{SOS1}, a relaxation technique for polynomial systems. However, such formulations are sensitive to numerical errors and scale poorly to high dimensions---both in theory~\cite{SOS_survey} and in practice~\cite{Rav:19}. Moreover, SOS methods tend to underestimate the actual funnel size~\cite{neural,neural_funnels}. 

To alleviate these~drawbacks of SOS methods, we propose the use of falsifiers based on numerical optimization to compute funnel candidates directly. We leave potential formal verification to the end. This allows for an efficient funnel optimization loop, since the dimensions of subsequent nonlinear programming (NLP) problems are the same as the dimension of the system and do not further increase as it is the case of SOS methods, which increases at least quadratically  in the problem dimension~\cite{SOS_survey}. Our computational experiments show that without the~verification part, the~falsifiers still provide quite accurate estimates of control funnels. As an additional advantage we note that the method is also applicable to non-polynomial systems that SOS-based methods cannot handle directly (i.e., they need the non-polynomial dynamics to be approximated by a polynomial one).

In further related work~\cite{rel_work2,rel_work3}, funnels are computed using linearization of system dynamics with conservative estimates of nonlinear effects. Unfortunately, both publications lack a direct comparison with methods based on SOS on the same benchmark problem which makes an estimation of their performance in terms of scalability and conservativeness difficult. Verification methods for ODEs~\cite{nedialkov} and hybrid systems~\cite{flowstar} insist on formal verification of the results, but compute an overapproximation of all trajectories from a given initial set instead of funnels. This means that system trajectories that start within a computed overapproximation but \emph{not} from the initial set are not guaranteed to stay in the overapproximation.
Concerning SOS relaxations, it would also be possible to uses alternatives, namely DSOS and SDSOS~\cite{rel_work1}. This improves scalability but results in more conservative solutions which would result in smaller funnels.

The~structure of the~paper is as follows. In~Section~\ref{sec_problem}, we state the~precise problem. In~Section \ref{sec_funnel}, we review the problem of funnel construction and describe existing approaches based on SOS programming. In Section~\ref{sec_construction}, we introduce our algorithm and explain its implementation. In~Section~\ref{sec_examples}, we provide computational experiments. Section~\ref{sec_conclusion} concludes the paper.

\section{Problem Statement}
\label{sec_problem}

Consider a~system 
\begin{equation}
\label{system}
\dot{x} = F(x,t) 
\end{equation}
where $F\colon\mathbb{R}^n\times\mathbb{R}\mapsto\mathbb{R}^n$ is a~smooth function. We further assume that~system~\eqref{system} has a~unique solution for any initial point $x_0\in\mathbb{R}^n$ and time $t_0\geq 0$. We denote this solution by $\Sigma_{(x_0, t_0)}$, which~is a~function in $[t_0,\infty]\mapsto\mathbb{R}^n$. We will also simply write $\Sigma_{(x_0)}$ for $\Sigma_{(x_0, 0)}$.

Let  $x_0\in\mathbb{R}^n$ and $T>0$. In this paper, we consider the problem of computing funnels~\cite{SOS_funnels}. A funnel is a time-varying set of states $\mathcal{F}(t)$ for $0\leq t\leq T$ such that for all $t \in[0, T]$ and $x \in \mathcal{F}(t)$, the solution from $(x, t)$ stays in the funnel, i.e., $\Sigma_{(x, t)}(\tau) \in \mathcal{F}(\tau)$ for all $\tau\in [t, T]$. In addition, we require that the final part of a funnel $\mathcal{F}(T)$ is a subset of some chosen set of goal states $\mathcal{G}$. Finally, we also want a funnel as large in volume as possible.

To ease the construction of funnels, we reduce our attention to funnels $\mathcal{F}(t)$ that are constructed around some chosen system trajectory $\tilde{x}(t)  = \Sigma_{(x_0)}(t)$ for $t\in[0, T]$ that ends in $\mathcal{G}$~\cite{LQRtrees1,SOS_funnels}. These funnels can be described using a differentiable positive definite function $P(x,t)\colon \mathbb{R}^n\times[0, T]\mapsto\mathbb{R}$, and a~differentiable\footnote{actually, continuous and right differentiable suffices} real function $\rho(t): [0, T]\mapsto (0,+\infty)$ as sublevel sets $\mathcal{F}(t) = \{x\in\mathbb{R}^n \mid\ P(x-\tilde{x}(t),t)\leq \rho(t)\}$. Hence, funnel construction reduces to construction of functions $P(x,t)$ and $\rho(t)$ such that $\mathcal{F}(t)$ forms a funnel as large in volume as possible. 

Simplifying the problem even further, we will assume that the shape $P(x,t)$ is provided beforehand e.g. as an ellipsoid given by a solution of the corresponding Lyapunov~\cite{Khalil} or Riccati equation~\cite{Lib:11}, or an optimized one for the linearized model using matrix inequalities~\cite{boyd:94}. Consequently, all that remains is to determine an optimal $\rho(t)$ wrt. fixed $P(x,t)$.

\section{Funnel construction}
\label{sec_funnel}
In this section, we will shortly review general funnel construction and an SOS-based variant~\cite{LQRtrees1,SOS_funnels}. We assume a set  $\mathcal{F}(t) = \{x\in\mathbb{R}^n \mid\ P(x-\tilde{x}(t),t)\leq \rho(t)\}$ as described in the previous section. First, we explore conditions on $P(x,t)$ and $\rho(t)$ that make $\mathcal{F}(t)$ a funnel. Let us define for each $t\in[0, T]$ sublevel sets $L^\leq_{P, \rho}(t) \equiv \{x\in\mathbb{R}^n \mid\ P(x,t)\leq \rho(t)\}$, and level sets $L_{P, \rho}(t) \equiv \{x\in\mathbb{R}^n \mid\ P(x,t) = \rho(t)\}$. Hence, we can write $\mathcal{F}(t) = L^\leq_{P, \rho}(t) \oplus \tilde{x}(t)$, where $\oplus$ denotes Minkowski addition. 

Assume that $P(x,t)$ and $\rho(t)$ are chosen in such a~way that the final sublevel set is a subset of a target set, i.e., $L^\leq_{P, \rho}(T) \oplus \tilde{x}(T) \subseteq \mathcal{G}$. Moreover, assume that for all $t\in[0, T]$ and all $x \in L_{P, \rho}(t) \oplus  \tilde{x}(t)$ the value of $P$ decreases faster or increases slower than $\rho(t)$ along the system dynamics, that is
\begin{equation}
\label{main_condition}
\dot{P}(x - \tilde{x}(t),t) < \dot{\rho}(t),
\end{equation}
where $\dot{P}(x,t) = \nabla_x P(x,t)^T \dot{x}+ \frac{\partial}{\partial t}P(x,t).$ Due to this requirement, for all $t\in[0, T]$, all states $x$ in $L^\leq_{P, \rho}(t) \oplus \tilde{x}(t)$  stay  in $L^\leq_{P, \rho}(t) + \tilde{x}(t)$, for all $\tau\in [t, T]$. Consequently, the set $L^\leq_{P, \rho}(t) \oplus \tilde{x}(t)$ forms a funnel.

As we mentioned in the previous section, we reduce our attention to the case in which $P(x,t)$ is fixed beforehand. An~often suitable candidate for $P(x,t)$ is a~solution of the Lyapunov or Riccati equation. A candidate then has quadratic form  $P(x,t) = x^TS(t)x,$ where $S(t)$ is a solution to a respective equation. Still, even if we fix $P(x,t)$, we still need to determine $\rho(t)$. Additionally, we would like $\rho(t)$ to be chosen in such a~way that the~sublevel sets are as large as possible.

In previous work~\cite{LQRtrees1,SOS_funnels}, $\rho(t)$ is parametrized piecewise-linearly and the parameters are optimized using a line-search approach. In each iteration, $\rho(t)$ is verified using SOS programming. Moreover, computation of $\rho(t)$ can be approximated by performing it in finitely many time samples~\cite{LQRtrees1,SOS_funnels}. This partially alleviates the needed computational burden due to ignoring polynomial dependence in $t$. However, certain care must be taken with choosing  time samples to obtain a reasonable approximation of a funnel as we will see later in  Example 1 of our computational experiments.

To be more specific, let us assume that both $P(x-\tilde{x}(t),t)$ and  $\dot{P}(x-\tilde{x}(t),t)$ are polynomials in $x$. Choose  time instants $0 = t_1 < t_2 < \cdots < t_N =T$, denote  $\rho(t_1), \ldots, \rho(t_N)$ as $\rho_1, \ldots, \rho_N$, and set $\rho(t) = \rho_i + \frac{\rho_{i+1}-\rho_i}{t_{i+1}-t_i}t$ for an interval $t\in[t_i,t_{i+1}].$ To optimize the volume of a discrete funnel,  a linear cost  $\sum_{i = 1}^N \rho_i$ is considered for optimization~\cite{SOS_funnels}. Hence, values $\rho_1, \ldots, \rho_N$ for which $\rho(t)$ meets \eqref{main_condition} in all time samples are found by solving a bilinear SOS program 
\begin{gather}
\max_{\rho_1,\ldots, \rho_{N}, \epsilon, \mu_1, \ldots \mu_N } \sum_{i = 1}^N \rho_i \nonumber\\
\text{subject to}\label{SOS}\\
\allowdisplaybreaks
L^\leq_{P, \rho}(T) + \tilde{x}(T) \subseteq \mathcal{G}\nonumber,\\
\begin{multlined}[\textwidth/3]
\varepsilon - \dot{P}(x-\tilde{x}(t_i),t_i) + \frac{\rho_{i+1}-\rho_i}{t_{i+1}-t_i} +\nonumber\\
+ \mu_i(x)(\rho_i - P(x-\tilde{x}(t_i),t_i)) \nonumber\\
\text{ is an SOS polynomial}\; \forall i = 1 , \ldots, N-1
\end{multlined}
\end{gather}
where $\varepsilon>0$ and $\mu_1, \ldots \mu_N$ are real polynomials. Note that the constraint in \eqref{SOS} is bilinear in $\mu$ and $\rho$, and thus an algorithm for solving \eqref{SOS} iteratively alternates between solving SOS program~\eqref{SOS} for multipliers $\mu$ and $\varepsilon$ with fixed $\rho$, and solving SOS program~\eqref{SOS} for $\rho$ with fixed $\mu$ and $\varepsilon$~\cite{LQRtrees1,SOS_funnels}. This also requires a valid initial funnel as described in~\cite{SOS_funnels}.  

SOS programming, while a~convex optimization problem, is computationally demanding, can encounter numerical problems, and scales poorly to high dimensions~\cite{SOS_survey, Rav:19}. In particular, an SOS polynomial constraint in \eqref{SOS} can be reformulated as~\cite{SDP1} 
\begin{multline}
\label{SOScon}
\varepsilon - \dot{P}(x-\tilde{x}(t_i),t_i) + \frac{\rho_{i+1}-\rho_i}{t_{i+1}-t_i} + \\
+\mu_i(x)(\rho_i - P(x-\tilde{x}(t_i),t_i)) = z(x)^T Q z(x), 
\end{multline}
where $z(x)$ is a vector of monomials up to degree $d$, and $Q$ is an unknown semidefinite matrix with ${n+d\choose d} \times {n+d\choose d} \approx n^{2d}$ elements provided that polynomial on the left hand side is of degree $2d$~\cite{SOS_survey}. Since polynomials are equal only if their coefficients are equal, constraint \eqref{SOScon} can be replaced with  ${n+d\choose d} \times {n+d\choose d}$ equalities (\emph{coefficient matching conditions}~\cite{SOS_survey}) and one semidefinite matrix constraint $Q \geq 0$. Hence, the states $x$ are removed from the optimization, but a new semidefinite matrix variable $Q$ is introduced, which causes the aforementioned scalability issues in SOS programming~\cite{SOS_survey}.

Moreover, the approach requires repeated solving of~\eqref{SOS} to perform optimization over $\mu$ and $\rho$. Also note that a resulting value of $\rho$ may not be optimal, since the problem~\eqref{SOS} is bilinear (i.e., non-convex), and the transformation to SOS is a relaxation technique. A final slight drawback of SOS relaxation is that system dynamics $F$ \eqref{system} and $P$ must be polynomials.

\section{Constructing $\rho$ using numerical optimization}
\label{sec_construction}

\begin{algorithm}
  \caption{Funnel Synthesis}
  \label{alg}
  \begin{description}
\item[In:] A~system $\dot{x} = F(x,t)$, a goal region $\mathcal{G}$, a reference trajectory $\tilde{x}(t)$ for $t\in[0,T]$ with $\tilde{x}(T)\in\mathcal{G}$, positive definite function $P(x,t)$, time samples
$$ 0 = t_1 < \cdots < t_N = T,$$
and sampling for each interval $[t_k, t_{k+1}]$.
\item[Out:] Funnel $\mathcal{F}(t), t\in [0, T]$
  \end{description}
  \begin{enumerate}
  \item Let $\rho_N$ be s.t. $L^\leq_{P(T), \rho_{N}} + \tilde{x}(T) \subseteq \mathcal{G}$
\item For $k:= N-1, N-2, \ldots, 1$
\begin{enumerate}
\item Put $\rho_k \coloneqq c\rho_{k+1}$
\item Repeat until $\tau_1$ subsequent iterations do not change $\rho_k$
  \begin{itemize}
  \item   Solve~\eqref{iter_NLP1} from a random initial point. 
  \item If the solution evolves outside of $L^\leq_{P(t_{k+1}), \rho_{k+1}} + \tilde{x}(t_{k+1})$, find a solution $x'$ to~\eqref{iter_NLP2} and put $\rho_k \coloneqq \gamma_1 P(x',t_k)$.
  \end{itemize}
\item \label{step:deriv_check} Repeat until $\tau_2$ subsequent iterations do not change $\rho_k$
  \begin{itemize}
  \item   Solve~\eqref{iter_NLP3} for the respective sampling from random initial points. 
  \item If the solution does not meet \eqref{discrete_condition}, put $\rho_k \coloneqq \gamma_2 \rho_k$ 
  \end{itemize}
\end{enumerate}
\item return the funnel $\mathcal{F}(t):= \{ P(x,t) \leq \rho^I(t) \}$, where $\rho^I(t)$ is a piece-wise linear interpolation between the samples $\rho_1, \ldots, \rho_N$
\end{enumerate}
\end{algorithm}

In this section, we describe a funnel computation algorithm that avoids the use of costly SOS programming. We propose the use of falsifiers based on numerical optimization to solve the optimization of $\rho$ and to leave potential formal verification to the end.

Our algorithm samples the constructed funnels in time and proceeds backwards.  Let us choose time instants $0= t_1 < t_2 < \cdots < t_N =T$ and find for a given $P(T)$ a value~$\rho_N$, such that $L^\leq_{P(t_N), \rho_N} \oplus \tilde{x}(T) \subseteq \mathcal{G}$ is as large as possible. This NLP problem can be solved for $P(T)$ and $\mathcal{G}$ quadratic using semidefinite programming. Next, we compute  the samples $\rho(t_1), \ldots, \rho(t_{N-1})$ which we denote as  $\rho_1, \ldots, \rho_{N-1}$. Finally, we assume an interpolation between the samples and check condition \eqref{main_condition} for the interpolated funnel.

Three NLPs are to be solved for each time sample. The first two, NLPs~\eqref{iter_NLP1} and~\eqref{iter_NLP2}, are used to provide the time sampled optimal funnel (in terms of volume). The final one~\eqref{iter_NLP3} checks condition~\eqref{main_condition} that would be used for formal verification of the interpolated funnel. The algorithm shrinks the funnel, if any counterexample to condition~\eqref{main_condition} is found, or accepts the sampled value, if it does not.

Let us describe the algorithm more closely. Assume that we already determined the~optimal value of $\rho_{k+1}$. To determine the~optimal value of $\rho_k$, consider the NLP that seeks a point with smallest possible value $\rho_k$ for which the system leaves $L^\leq_{P(t_{k+1}), \rho_{k+1}} \oplus \tilde{x}(t_{k+1})$ after evolving from~$t_k$ to~$t_{k+1}$:
\begin{gather}
\min_{x} P(x-\tilde{x}(t_k),t_k)\nonumber\\
\text{subject to}\label{iter_NLP2}\\
P(\Sigma_{(x,t_k)}(t_{k+1})-\tilde{x}(t_{k+1}), t_{k+1})\geq \rho_{k+1}\nonumber
\end{gather}

NLP~\eqref{iter_NLP2} is non-convex, thus a local NLP solver can solve this NLP only approximately. Therefore, for reliably accepting a certain value $\rho_k$, more needs to be done. The first step to do so is another NLP
\begin{gather}
\max_{x} P(\Sigma_{(x,t_k)}(t_{k+1}))\nonumber\\
\text{subject to}\label{iter_NLP1}\\
P(x-\tilde{x}(t_k),t_k)\leq \rho_k, \nonumber
\end{gather}
that checks whether the current estimate $\rho_k$ results in a counterexample, a state that evolves outside of $L^\leq_{P(t_{k+1}), \rho_{k+1}} \oplus \tilde{x}(t_{k+1})$. If the found optimum is bigger than $\rho_{k+1}$, we found a counter-example, and hence we solve NLP~\eqref{iter_NLP2}, using the solution to NLP~\eqref{iter_NLP1} as an initial feasible estimate. This solution gives us a new, smaller estimate for $\rho_k$. If the found optimum is not bigger than $\rho_{k+1}$, we cannot make a definite conclusion, since NLP~\eqref{iter_NLP1} is again non-convex. Hence,  we increase the trust in the current estimate by repeatedly solving  NLP~\eqref{iter_NLP1} from random initial points until no further counter-example is found within a certain number $\tau_1$ of subsequent iterations. 

The use of NLP~\eqref{iter_NLP1} has two major advantages over only iterating NLP~\eqref{iter_NLP2} from random initial points. First, NLP~\eqref{iter_NLP1} directly checks for the existence of a counter-example, making it more efficient for this purpose, in our experience. And second, the result of NLP~\eqref{iter_NLP1} provides a much more useful starting point for NLP~\eqref{iter_NLP2} than random starting points.

To enforce termination of the loop between NLPs~\eqref{iter_NLP1} and~\eqref{iter_NLP2}, we update $\rho_k$ as $\gamma_1 P(x',t_k)$, where $x'$ is the found numerical solution of~\eqref{iter_NLP2} and $0<\gamma_1< 1$.  The loop must terminate after finitely many iterations, since  there must be a $\rho_k$ small enough such that no counterexample exists due to continuity of solutions of ordinary differential equations wrt. their initial conditions~\cite{Khalil} and the fact that $P(x,t)$ is positive definite. It should also be noted that, in general, we do not have $\Sigma_{(x,t)}$ available in explicit form, and hence we must approximate it using numerical integration.

After the end of the iteration between NLPs~\eqref{iter_NLP1} and~\eqref{iter_NLP2}, we try to extend the funnel from $t_k$ to the whole time interval $[t_k, t_{k+1}]$. For this we use linear interpolation between $\rho_k$ and $\rho_{k+1}$. Based on this, we would have to check condition~\eqref{main_condition} for all $t\in[t_k, t_{k+1}]$ and all $x \in L_{P, \rho}(t) + \tilde{x}(t)$. However, as mentioned in \cite{LQRtrees1,SOS_funnels}, this is not convenient to check due to dependency on time $t$. It is computationally far more efficient (for both SOS relaxation and our presented approach) to simply sample the time interval $[t_{k}, t_{k+1}]$ and to check the~condition discretely. Moreover, continuity arguments show~\cite{SOS_funnels} that provided that sampling is fine enough, no counterexamples to condition~\eqref{main_condition} may exist. 

Assume a~sampling $t_{k,1}, \ldots, t_{k,M}$ of interval $[t_{k}, t_{k+1}]$, and denote by $\rho^I$ 
the linear interpolation between $\rho_k$ and $\rho_{k+1}$ used in \cite{LQRtrees1,SOS_funnels}. Then $\dot{\rho}^I(t)= \frac{\rho_{k+1}- \rho_k}{t_{k+1}-t_k}$, and  for $j = 1, \ldots, M$, and all $x \in L_{P, \rho^I}(t_{k,j}) \oplus \tilde{x}(t_{k,j})$, we must ensure 
\begin{equation}
\label{discrete_condition}
\dot{P}(x - \tilde{x}(t_{k,j}),t_{k,j}) \leq \dot{\rho}^I(t_{k,j}).
\end{equation}
Since $\dot{P}$ tends to zero for $x \rightarrow\tilde{x}$, we can guarantee that these conditions can be met by choosing $\rho_k$ small enough. Thus again the~resulting algorithm will succeed in finitely many iterations. However, we need to employ a line-search strategy on $\rho_k$ to obtain an optimal funnel that meets \eqref{discrete_condition}. 

We can check condition~\eqref{discrete_condition} by numerical optimization, namely by solving an~NLP
\begin{gather}
\max_{x} \dot{P}(x-\tilde{x}(t_{k,j}),t_{k,j}) \nonumber\\
\text{subject to}\label{iter_NLP3}\\
P(x-\tilde{x}(t_{k,j}),t_{k,j}) = \rho^I(t_{k,j})\nonumber
\end{gather}
which is again a~non-convex problem. Again we ensure reliability of the check by solving the NLP repeatedly from random initial points until no more counter-examples appear. If a counter-example is found, we reduce $\rho_k$ using a multiplier $0<\gamma_2<1$.  

Algorithm~\ref{alg} summarizes the whole algorithm. We estimate the initial value of $\rho_k$ as $c\rho_{k+1}$, where $\rho_{k+1}$ is the previous computed value. Note that $c$ should be chosen large enough to ensure that the first estimate always contains a counterexample, and thus the first estimate is always an upper bound of the optimal funnel size.

\section{Computational Experiments}
\label{sec_examples}

In this section, we discuss the results of computational experiments using the method from the previous section. The implementation was done in MATLAB R2017b and ran on a~PC with Intel Core i7-10700K, 3.8GHz and 32GB of RAM. We will do a comparison between the method described in the previous section and the SOS method described in~\cite{SOS_funnels}. The NLP solver used for our method and for generation of reference trajectories was implemented in CasADi~\cite{Andersson2018} with internal NLP solver ipopt~\cite{Ipopt}. The SOS method was implemented in Yalmip~\cite{Lofberg2004} with internal SDP solver Mosek~\cite{mosek}.

\begin{figure}
\begin{subfigure}{.45\textwidth}
\centering
\includegraphics[width=.95\linewidth]{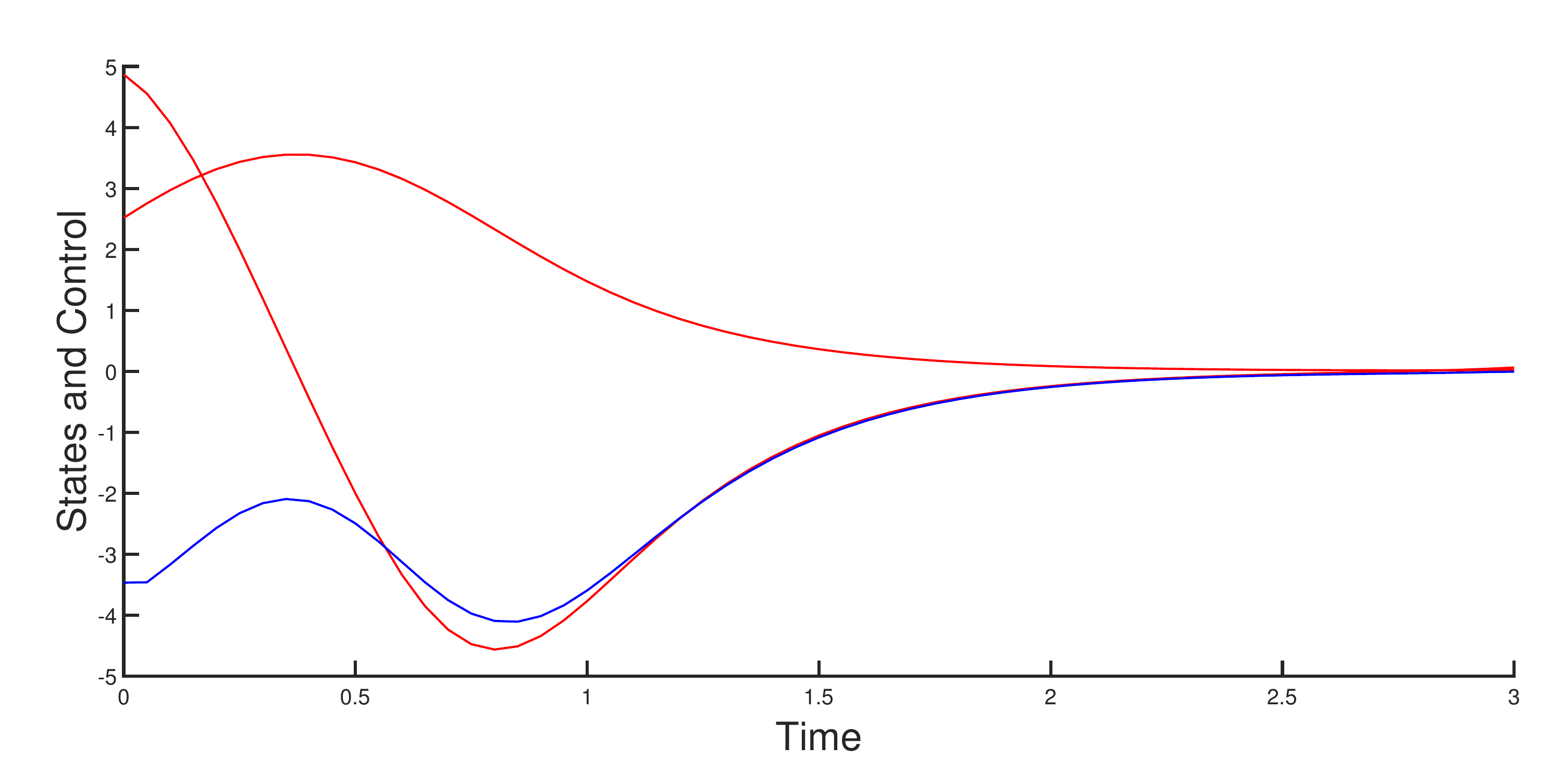}
\caption{Pendulum: states (red) and control (blue)}
\label{traja}
\end{subfigure}
\begin{subfigure}{.45\textwidth}
\centering
\includegraphics[width=.95\linewidth]{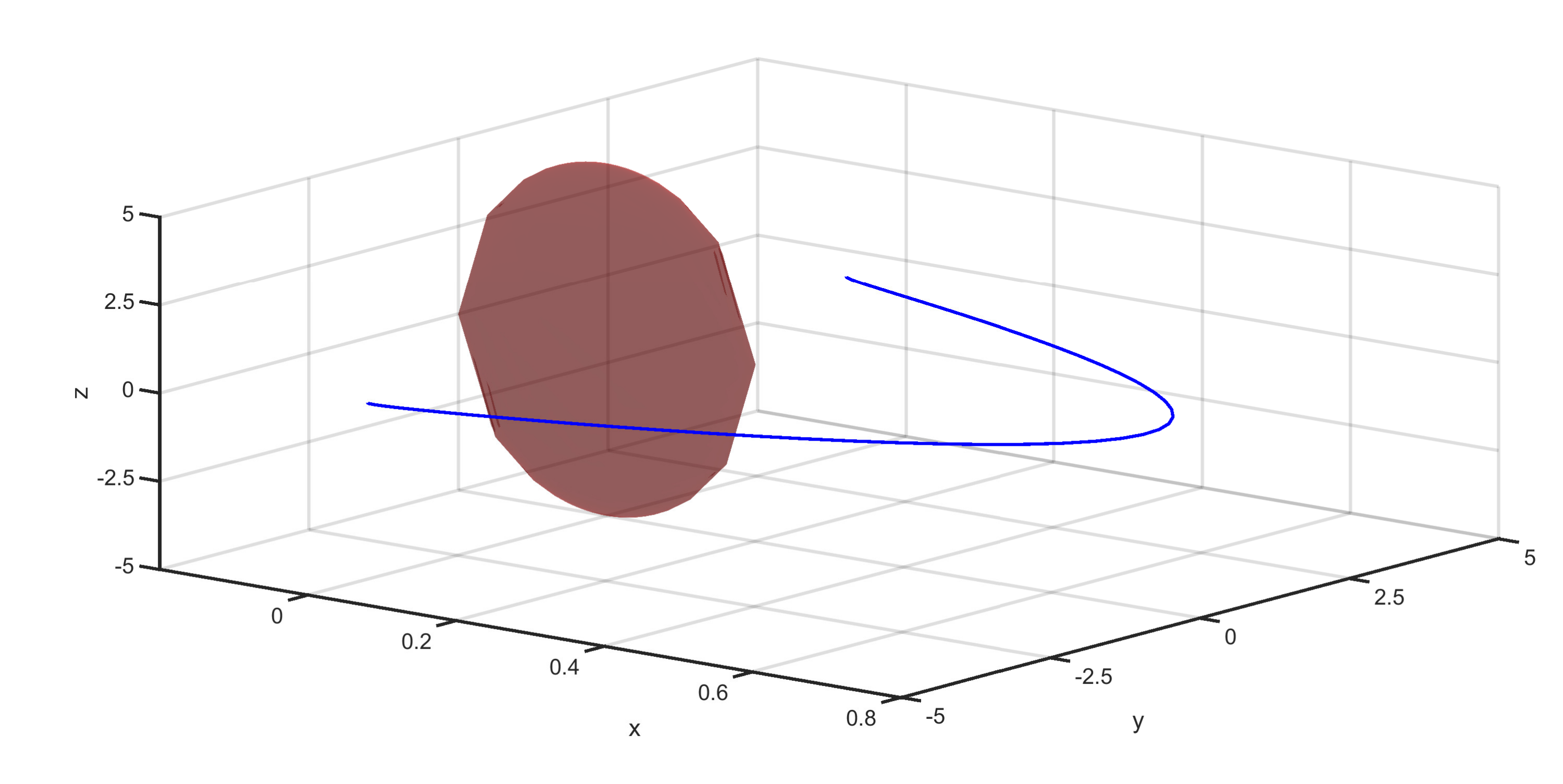}
\caption{Quadcopter: trajectory (blue) and obstacle (red)}
\label{trajb}
\end{subfigure}
\caption{Reference trajectories}
\end{figure}

\subsection{Example 1: Inverted pendulum}

We start with a simple two dimensional problem, an inverted pendulum, and continue with more involved examples later. The~dynamics of the~inverted pendulum are
\begin{equation} 
\label{pendulum}
\ddot{\theta} = \frac{g}{l}\sin\theta - \frac{b\dot{\theta}}{ml^2}  + \frac{u}{ml^2},
\end{equation}
where we set $m = 1, l = 0.5, g = 9.81, b = 0.1.$ Assume the task of steering an inverted pendulum to its unstable equilibrium $\bar{x} = 0$. First, we computed a stabilizing  reference trajectory of length $T = 3$ with step $0.05$ using CasADi, see Figure~\ref{traja}. Next, we constructed an LQR tracking controller for the interpolated reference (piecewise cubic in states and piecewise linear in control) by solving the Riccati equation for $R = Q = I$ with a final value of a cost-to-go matrix $S(T) = Q$ using the RKF45 integrator with a maximum step $0.005$ and used again cubic interpolation, and hence we obtained matrices $S(t)$ for $t\in [0, T]$. We set $\mathcal{G} = \{x\in\mathbb{R}^2 \mid (x-\tilde{x}(T))^T(x-\tilde{x}(T)) \leq 0.0025\}$ as our target set. 

For the SOS method, we set an initial feasible funnel as $\rho(t) = 0.0025\,\mathrm{exp}\left(0.6\cdot\frac{T-t}{T}\right)$ using the template from~\cite{SOS_funnels}. Additionally, we approximated the non-polynomial dynamics with cubic Taylor polynomials and set $\mu$ to be quadratic. We did not use polynomials of higher degree due to numerical problems encountered by the SDP solver. We terminated the SOS algorithm, if the volume of the funnel increased less than $0.1\%$ between two subsequent iterations or if the solver failed due to numerical errors. In our method, we set $\gamma_1 = 0.9999$, and $\gamma_2 = 0.999$, and the iteration bounds $\tau_1=10$ and $\tau_2=30$ and computed funnels for both the original and the polynomial model. We used just one sample for each interval ($t_{k+1}$ for the interval $[t_k, t_{k+1}]$) in evaluating derivatives for both methods.

\begin{figure}
\centering
\includegraphics[width=\linewidth]{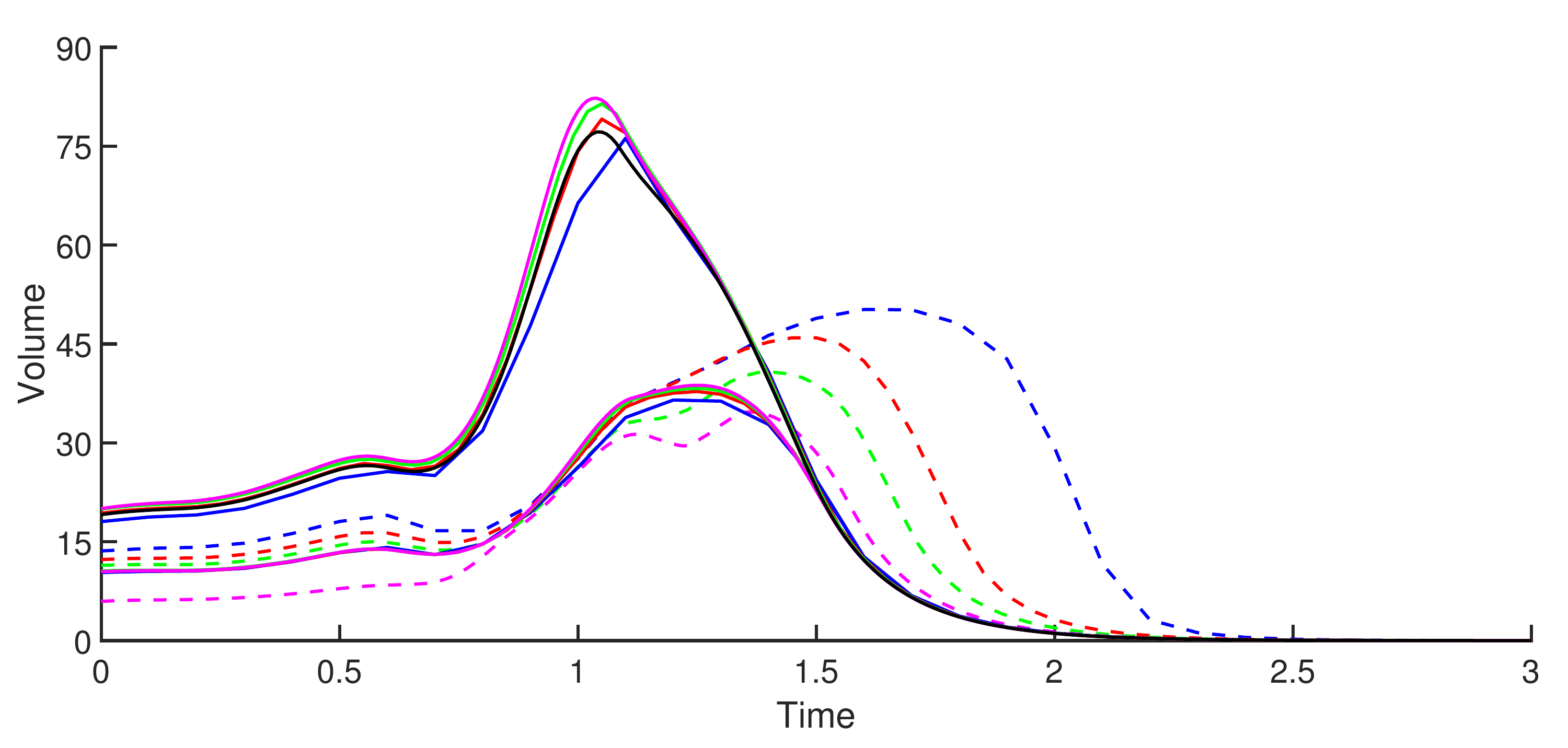}
\caption{Inverted pendulum: Funnels for falsifier based (full line) and SOS based (dashed line) methods for time steps $h = 0.1, 0.05, 0.03, 0.01$ (blue, red, green, purple).}
\label{pendulum_funnels}
\end{figure}

The results can be seen in Figure~\ref{pendulum_funnels} and Table \ref{result_table}. The SOS method is faster and SOS funnels are larger for the last time intervals. However, the computed funnels are incorrect in the sense that they do not relate to an actual funnel as we have tested via numerical simulations of the polynomial system. The reason for this problem is the fact that the derivatives are checked too sparsely for this example. Falsifier based funnels do not suffer from this in this example since funnel sizes are also estimated from above by numerical integration, not just by derivatives alone. Hence, sparser sampling is needed to obtain an accurate estimate of the actual funnel in comparison to the SOS method. We also tried SOS to verify the computed funnels and while the solver did not falsify them, the solver reported some solutions to not to be reliable. In terms of volume, the cubic Taylor approximation underestimates the funnel size and this example would require a significantly higher (about 7 according to our method) degree polynomial to accurately describe the actual funnel.

\begin{table}
\centering
\begin{tabular}{c || c|c || c|c|c }
\multicolumn{1}{c||}{Pend} & \multicolumn{2}{c||}{Fals (poly/orig)} & \multicolumn{3}{c}{SOS} \\
\hline
$h$ & $t$ & $\mathrm{Vol}$ & iter & $t$ & $\mathrm{Vol}$ \\
\hline
0.10 & 17.2/15.7 & 34.2/59.8 & 18 &  3.8 & 63.5 \\
0.05 & 18.3/17.5 & 34.6/62.4 & 20 & 7.5  & 48.2 \\
0.03 & 24.7/24.5 & 34.9/63.9 & 13 & 7.6 & 40.5 \\
0.01 & 70.8/68.1 & 35.0/64.4 & 8 & 14.0 & 30.3 \\
\hline
\multicolumn{1}{c||}{Quad} & \multicolumn{2}{c||}{Fals (poly/orig)} & \multicolumn{3}{c}{SOS/SOS+fals} \\
\hline
$h$ & $t$ & $\mathrm{Vol}$ & iter & $t$ & $\mathrm{Vol}$  \\
\hline
0.10 & 50/24 & 9.7/9.7 & 7/3 & \multicolumn{1}{c|}{538/158}     & \multicolumn{1}{c}{9.1/9.7} \\
0.05 & 44/92 & 12.2/12.1 & 12/3 & \multicolumn{1}{c|}{1784/313} & \multicolumn{1}{c}{12.0/12.1} \\
0.03 & 76/206 & 13.6/13.5 & 7/2 & \multicolumn{1}{c|}{1536/251}  & \multicolumn{1}{c}{13.3/13.3}\\
\end{tabular}
\caption{Quadcopter and pendulum: results for both methods, time required $t$, and volume $\mathrm{Vol}$ of the funnel  (falsifier based method) and number of iterations, and time required in the SPD solver Mosek $t$, and volume $\mathrm{Vol}$ of the funnel  (SOS based method). Volume of funnels for the quadcopter example is given in $10^{-8}$ units.} 
\label{result_table}
\end{table}

\subsection{Example 2: Quadcopter}

Let us consider a~twelve dimensional problem. We assume the~quadcopter model (2.30) -- (2.35) in~\cite{garcia2013modeling} for unit mass and the obstacle avoidance manoeuvre from Figure~\ref{trajb} computed using CasADi. We again constructed an LQR tracking controller for the interpolated discrete trajectory (piecewise cubic in states and piecewise linear in control) by solving the Riccati equation for $Q = 10I$ and $R = I$ with final value of a cost-to-go matrix $S(T) = Q$ using the RKF45 integrator with a maximum step $0.001$ and again used cubic interpolation, and hence we obtained matrices $S(t)$ for $t\in [0, T]$. We set $\mathcal{G} = \{x\in\mathbb{R}^{12} \mid (x-\tilde{x}(T))^T(x-\tilde{x}(T)) \leq 0.1\}$ as our target set.

For the SOS method, we set an initial feasible funnel as $\rho(t) = 0.1\,\mathrm{exp}\left(-1.94\cdot\frac{T-t}{T}\right)$. We approximated non-polynomial dynamics with cubic Taylor polynomials, that accurately approximates the original dynamics,  and set $\mu$ to be quadratic. We terminated the SOS algorithm, if the volume of the funnel increased less than $0.1\%$ between two subsequent iterations. In our method, we again set $\gamma_1 = 0.9999$, $\gamma_2 = 0.999$, $\tau_1=10$ and $\tau_2=30$. We again considered just one sample for each interval ($t_{k+1}$ for the interval $[t_k, t_{k+1}]$) in evaluating derivatives for both methods.

\begin{figure}
\centering
\includegraphics[width=\linewidth]{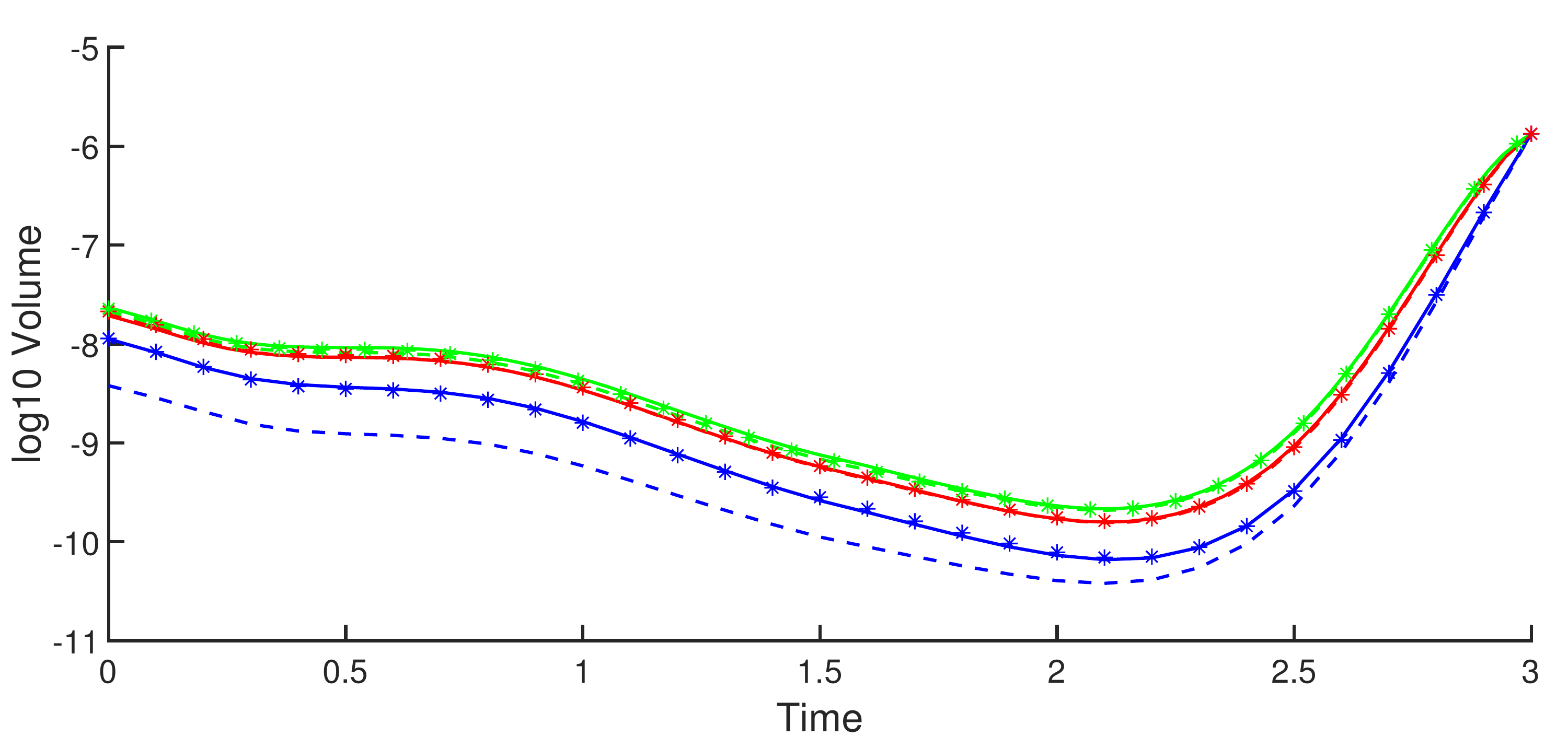}
\caption{Quadcopter: Funnels for falsifier based (full line), and SOS based (dashed line), and SOS + falsifier based (stars) methods for time steps $h = 0.1, 0.05, 0.03$ (blue, red, green).}
\label{quadcopter_funnels}
\end{figure}

The results can be seen in Figure \ref{quadcopter_funnels} and Table~\ref{result_table}. Notice that, the falsifier based and SOS method actually (except for $h = 0.1$) provided very similar solutions which demonstrates the accuracy of our method.  However, the SOS method is significantly slower here ($20\times$ -- $50\times$). This shows lower of scalability of SOS programming in the problem dimension~\cite{SOS_survey}. But we can exploit the similarity of both solutions to significantly speed up the SOS method by combining both methods initializing
the SOS method by the result of ours. To illustrate this, we ran the SOS method with the initial funnel $\mathrm{exp}\left(-0.15\cdot\frac{T-t}{T}\right)\rho(t)$, where $\rho(t)$ was the funnel computed by our method, see Table~\ref{result_table}.

\subsection{Example 3: Pendulum revisited}

Let us return to a pendulum example, where we explore our method on problems of higher dimensions parametric in $n$. We assume a model of an $n$-link pendulum with $g = 9.81$ and we set the other parameters (all weights and lengths) to $1$. The derivation of equations of motion can be found in \cite{nlink}. The equation can be written in manipulator form
\begin{equation}
M(\theta,\dot{\theta})\ddot{\theta} + G(\theta,\dot{\theta}) = u,
\end{equation}
where we assume that $u\in\mathbb{R}^n$ is a control input. Next, we construct a nonlinear stabilizing controller for $n$-link pendulum 
\begin{equation*}
u(\theta,\dot{\theta}) = \left[ I, \; M(\theta,\dot{\theta})\right]^TK\left(\left[\theta,\dot{\theta}\right]^T - \left(\frac{\pi}{2}, 0, \ldots 0 \right)^T\right),
\end{equation*}
where $I$ is an identity matrix and $K$ is a gain matrix of the LQR controller based on a slightly simpler model $\ddot{\theta} + M(\theta,\dot{\theta})^{-1}G(\theta,\dot{\theta}) + u = 0$
linearized around the pendulum-upwards equilibrium $\left(\frac{\pi}{2}, 0, \ldots 0 \right)$.

\begin{table}
\centering
\begin{tabular}{r || S|| S || S || r || S|| S || S }
& \multicolumn{1}{c||}{Original } & \multicolumn{1}{c||}{Linear} & \multicolumn{1}{c||}{SOS} &  & \multicolumn{1}{c||}{Original } & \multicolumn{1}{c||}{Linear} & \multicolumn{1}{c}{SOS}\\
\hline
$n$ & $t$ & $t$ & $t$ & $n$ & $t$ & $t$ & $t$\\
\hline
1 & 13.3 & 13.6 & 0.3 & 11 & 274.3 & 30.8 & 12378.2 \\
2 & 17.5 & 16.9 & 0.6 & 12 & 379.9 & 33.0 & {}\\
3 & 19.7 & 18.2 & 1.8 & 13 & 543.1 & 34.4 & {}\\
4 & 26.5 & 20.4 & 7.2 & 14 & 752.4 & 36.4 & {}\\
5 & 40.5 & 23.9 & 31.1 & 15 & 1044.6 & 38.8 & {}\\
6 & 50.1 & 23.8 & 87.4 & 16 & 1326.5 & 41.3 & {}\\
7 & 66.6 & 23.9 & 353.5 & 17 & 1819.2 & 45.4 & {}\\
8 & 90.9 & 25.4 & 840.5 & 18 & 2366.7 & 49.6 & {}\\
9 & 128.1 & 26.8 & 2302.3 & 19 & 3046.4 & 50.6 & {}\\
10 & 185.2 & 28.7 & 5635.9 & 20 & 3990.7 & 53.4 & {}\\
\end{tabular}
\caption{$n$-link pendulum:} time required $t$ of the first iteration for the SOS method for the linearized model, and the time required $t$ for the falsifier based method for the original model and its linearization
\label{nlink_table}
\end{table}

We compute funnels for stabilization of the $n$-link pendulum with the derived controller for $Q_n = 10nI$ and $R = 1$. We set a target $\mathcal{G}_n = \left\{\left[\theta,\dot{\theta}\right]^T\in\mathbb{R}^{2n} \mid \Delta ^T S_n \Delta \leq \rho_n \right\}$, where $\Delta = \left(\left[\theta,\dot{\theta}\right]^T-\left(\frac{\pi}{2}, 0, \ldots 0 \right)^T\right)$, and $S_n$ is a cost-to-go matrix of the LQR controller, and where $\rho_n$ is chosen in such a way that the volume of $\mathcal{G}_n$ is the same as the volume of a hypersphere with radius of $r^2 = 0.025$ in $2n$ dimensions. Notice that the whole problem is time-invariant since the chosen system trajectory around which we will construct a funnel is constant $\tilde{x}(t) = \left(\frac{\pi}{2}, 0, \ldots 0 \right)^T$ as well as the shape $S(t) = S_n.$

We again set  for our method $\gamma_1 = 0.9999$, $\gamma_2 = 0.999$. Moreover, we used $\tau_1=  10$ and $\tau_2=50$. And we again considered just one sample for each interval ($t_{k+1}$ for the interval $[t_k, t_{k+1}]$) in evaluating derivatives.

We computed funnels of length $T = 1$ with a time step $0.025$ for $n = 1, \ldots, 20,$, i.e., for state dimensions up to $40$. For a comparison, we tested our method on the original $n$-link pendulum model and  its linearized model. The results can be seen in Table \ref{nlink_table}. As can be seen from the results, the funnels were successfully computed for all $n$. However, the required computational time increases steadily for the original model, approximately by factor of one third for each new link added. This is mostly caused by the fact that system dynamics become more and more complex with each link added, which steadily increases computational time required for evaluation of system dynamics and its first and second order derivatives. 

It should be noted however that the computational time remained much more reasonable for the linearized model where this increase in complexity naturally does not occur. This shows that our method can work reasonably well even in high dimensions provided that a model dynamics are not too complicated. To illustrate scalability of the SOS method, we computed one iteration of the SOS method on the lineariazed dynamics with $\mu$ set as quadratic. As can bee seen in Table~\ref{nlink_table}, the computational time increases dramatically and becomes impractical with about $9$ links.

For the linearized model, we can compare the computed values of $\rho$  with the true optimal values. These can be computed directly for linear systems with an ellipsoidal funnel using the state transition matrix. Assume a linear system $\dot{x} = Ax$ and an ellipsoid $E = \{x \in\mathbb{R}^n \mid x^T Q^{-1} x \leq 1\}$. Using and affine mapping with the matrix $e^{At}$,  this ellipsoid transforms  into the ellipsoid 
$$E(t) = \left\{x \in\mathbb{R}^n \mid x^T \left(e^{At} Q e^{A^Tt}\right)^{-1} x \leq 1\right\}$$ after time $t$. 
Hence, the optimal value $\rho$ for a funnel with a given shape $S$ in time $t$ that ends in the ellipsoid $E$ is the  ellipsoid of maximal value $\rho$ for which $\{x \in\mathbb{R}^n \mid x^T S x \leq \rho \} \subseteq E(-t).$

\begin{table}
\centering
\begin{tabular}{r ||S|| S | S||  S}
& \multicolumn{1}{c||}{Original } & \multicolumn{1}{c}{Linear, DC} & \multicolumn{1}{c||}{Linear, no DC} & \multicolumn{1}{c}{Linear, optimal}\\
\hline
$n$ & \multicolumn{1}{c||}{$\rho(0)$} & \multicolumn{1}{c}{$\rho(0)$} & \multicolumn{1}{c||}{$\rho(0)$} & \multicolumn{1}{c}{$\rho(0)$}\\
\hline
1 & 18.48 & 15.48 & 15.48 & 15.54\\
2 & 6.24 & 6.13 & 6.13 & 6.16 \\
3 & 1.17 & 1.17 &  1.17 & 1.17  \\
4 & 1.26 & 1.26 & 1.31 & 1.32 \\
5 & 1.70 & 1.70 & 1.77 & 1.78 \\
6 & 1.79 & 1.79 & 1.94 & 1.95\\
7 & 1.96 & 1.97 & 2.13 & 2.14\\
8 & 2.05 & 2.06 & 2.32 & 2.34\\
9 & 2.24 & 2.26 & 2.55 & 2.56\\
10 & 2.36 & 2.38 & 2.79 & 2.81\\
11 & 2.57 & 2.59 & 3.04 & 3.06\\
12 & 2.77 & 2.80 & 3.30 & 3.32\\\
13 & 2.87 & 2.91 & 3.56 & 3.58\\
14 & 3.06 & 3.12 & 3.81 & 3.83\\
15 & 3.25 & 3.33 & 4.07 & 4.09\\
16 & 3.45 & 3.40 & 4.32 & 4.34\\
17 & 3.50 & 3.59 & 4.57 & 4.59\\
18 & 3.69 & 3.80 & 4.82 & 4.85\\
19 & 3.88 & 4.00 & 5.08 & 5.10\\
20 & 4.07 & 4.20 & 5.34 & 5.37\\
\end{tabular}
\caption{$n$-link pendulum:} value $\rho(0)$ for the original model and its linearization (with and without derivative check (DC)), and the optimal value of $\rho(0)$ for the linearized model computed via the state transition matrix
\label{nlink_table2}
\end{table}

We computed the optimal values of $\rho$ with a time step $0.025$ iteratively. The comparison of the resulting values of $\rho(0)$ can be seen in Table \ref{nlink_table2}. The values computed for the linearized model are slightly lower than the optimal ones. This underestimation of the funnels is largely caused by the derivative check \eqref{iter_NLP3} that assumes a piece-wise linear $\rho$, which the optimal $\rho$ is not. If the derivative check is skipped, our results and the optimal values are nearly identical.

\section{Conclusion}
\label{sec_conclusion}

In this paper, we presented an~algorithm that computes funnels along trajectories of systems of ordinary differential equations. Compared to related work based on SOS programming, in our computational experiments, the algorithm computed larger funnels in less time. The algorithm does not formally verify in itself, but its result can then be formally verified using a well-known palette of verification techniques that includes---in addition to SOS programming---computer algebra~\cite{Collins:91} or interval computation~\cite{Ratschan:02f}. In addition, these funnels could be used to initialize the SOS method as illustrated by Example 2.

\bibliographystyle{plain}
\bibliography{ddbibl}

\end{document}